\documentclass[conference,12pt,onecolumn]{IEEEtran}

\IEEEoverridecommandlockouts

\usepackage{amsmath,amssymb,amsfonts}
\usepackage{mathpazo}
\usepackage{subcaption}
\usepackage[ruled,vlined,linesnumbered]{algorithm2e}

\usepackage{graphicx}
\usepackage{xcolor}
\usepackage{lipsum}
\usepackage{pdfpages}
\usepackage{booktabs}

\def\BibTeX{{\rm B\kern-.05em{\sc i\kern-.025em b}\kern-.08em
    T\kern-.1667em\lower.7ex\hbox{E}\kern-.125emX}}

\newcommand\datasetOne{\xspace{ }CIFAR100\xspace}

\newcommand{\seclab}[1]{\label{sec:#1}}

\newcommand*{\B}[1]{\ifmmode\boldsymbol{#1}\else\textbold{#1}\fi}

                  {\unskip\nobreak\hskip 1em plus 1fil\nobreak%
                           \rule{2mm}{2mm}
                           \parfillskip=0pt%
                           \endtrivlist}

\begin{document}

\title{DSS: A \textbf{D}iverse \textbf{S}ample \textbf{S}election Method to Preserve Knowledge in Class-Incremental Learning}


\author{\IEEEauthorblockN{Sahil Nokhwal}
\IEEEauthorblockA{\textit{Dept. Computer Science} \\
\textit{University of Memphis}\\
Memphis, USA \\
nokhwal.official@gmail.com}
~\\
\and
\IEEEauthorblockN{Nirman Kumar}
\IEEEauthorblockA{\textit{Dept. Computer Science} \\
\textit{University of Memphis}\\
Memphis, USA \\
nkumar8@memphis.edu}
}

\maketitle

\begin{abstract}
	Rehearsal-based techniques are commonly used to mitigate catastrophic forgetting (CF) in Incremental learning (IL). The quality of the exemplars selected is important for this purpose and most methods do not ensure the appropriate diversity of the selected exemplars. We propose a new technique ``DSS'' -- Diverse Selection of Samples from the input data stream in the Class-incremental learning (CIL) setup under both disjoint and fuzzy task boundary scenarios. Our method outperforms state-of-the-art methods and is much simpler to understand and implement.
\end{abstract}


\section{Introduction} 
\seclab{intro}
In an incremental learning (IL) scenario, learning is done on a continuous stream of
data, rather than on a batch of data. The model is updated regularly (at so-called ``task boundaries'') and the learning algorithm cannot store all seen data due to space and computational limitations.  Catastrophic forgetting \cite{mccloskey1989catastrophic} is a problem with the model whereby it forgets what it has learned because old data is no longer accessible. One of the most common approaches to mitigate this problem is to store a representative sample of the previous data, known as exemplars.
In the task-incremental model of data, the model during inference has access to the task-ID by which it can refine its inference based on the knowledge of the set of classes in the task. In the class-incremental model, no such task-ID is present although the data stream is divided into tasks, and re-training happens at task boundaries. This paper refers to the class-incremental scenario and addresses both disjoint and fuzzy tasks. In disjoint tasks, classes in the tasks are disjoint and in the fuzzy model, some can repeat. Our contributions are as follows:
\begin{itemize}
    \item We propose an exemplar sampling technique(DSS) based on $k$-center clustering and is robust to outliers. 
    \item We demonstrate using extensive experiments on the \datasetOne \cite{cifar100} dataset, that both for disjoint and fuzzy10 (see Section~\ref{sec:prob_statement} for definition), the IL system using DSS outperforms the state-of-the-art methods.
\end{itemize}

\section{Related work}
\seclab{relatedWork}
We present a brief overview of existing research on offline methods. Refer to \cite{mai2022online} for detailed review.

\textbf{\emph{Offline Incremental Learning:}}  Conventional approaches to incremental learning, developed to counter CF, can be broadly categorized into three categories.

    

     \textbf{1) Regularization-based methods:} The model loss is adjusted to regulate retraining.
     There are two approaches for achieving this: 1) During model updates, the model parameters are regularized to penalize the more important weights, and 2) techniques that distill information from an old model developed on a previous input stream are used to regulate retraining. 
     Prominent techniques in this field include EWC \cite{kirkpatrick2017overcoming} and RTRA\cite{nokhwal2023rtra}. They determine which model parameters are more crucial by applying a greater penalty on the modifications of those parameters.

    \textbf{2) Parameter isolation methods:} Separate model parameters are used to deal with different tasks.
    Some parameters are constant across tasks and some are task-specific. The parameter isolation is expected to mitigate the forgetting of prior knowledge. The approach can be followed in both dynamic network architectures\cite{aljundi2017expert}, and fixed network topologies \cite{serra2018overcoming}.
    
    \textbf{3) Rehearsal-based methods:} Rehearsal-based methods either retain some input data (exemplars) or synthesize representative samples using GANs. During retraining, current task data, and previous exemplars or generated data are used \cite{lopez2017gradient,villa2023pivot,nokhwal2023pbes, nokhwal2023dss}. These methods also use a wide range of exemplar selection techniques \cite{aljundi2019task,shim2021online}.

\section{Problem formulation}
\label{sec:prob_statement}
The setup can be formulated as follows. The data for a CIL setup is received as a stream $(x_1, y_1), (x_2, y_2), \ldots, $ where
$y_i$ is the class label of data point $x_i$. The stream is conceptually divided into tasks.  Because we are in a class incremental scenario such task ID is not present at inference by definition. However, it can be present during training, though it is impractical to assume that such annotated data is present. Instead, our task boundaries, at which we retrain the model and select exemplars for the future, are defined by batching up data in some fashion - a batch of data will define a task. While we could implicitly keep track of a task ID, since it is not used, we do not do so.

In a disjoint CIL setting, the classes seen in each task are disjoint from each other. The number of classes that occur in each new task is the same. However, the number of data points in each class can vary - this is known as data imbalance. Therefore, the total number of data points can vary across tasks.

In a fuzzy$Z$ CIL setting where $Z$ denotes a percentage (usually a small number), we assume that each task has a certain fixed number of new classes that ``define'' the task, but it can also contain data from other classes and $Z\%$ of classes that define other tasks can show up in the current one. In this scenario, it is easy to see that one does not know which class defines a task and which is from others, and therefore the task boundaries are considered as blurry. The fuzzy or disjoint CIL configurations can be formulated as:
\begin{equation*} 
\begin{split}
   disjoint\_CIL \implies \bigcap ^ T_{t=0} = \Phi, \\
   fuzzyZ\%\_CIL \implies \bigcap ^ T_{t=0} \neq \Phi
\end{split}
\end{equation*}
Where, $t$ denotes any task, and $T$ is the total number of tasks.

 

The disjoint CIL configuration exacerbates CF because already-experienced classes are never exposed in subsequent tasks and thus makes this problem much harder to solve, whereas, in fuzzy$Z$ CIL, the division between tasks is unclear, such that $Z$\% of data from other (minor) classes will appear in every task and (100-$Z$)\% of data come from major classes. This is more general than disjoint CIL which can be considered as fuzzy0 CIL. 

\section{General architecture of rehearsal-based incremental learning}
\label{sec:general_arch}
\begin{figure*}[htbp!]
    \centering   
    \includegraphics[width=0.9\linewidth]{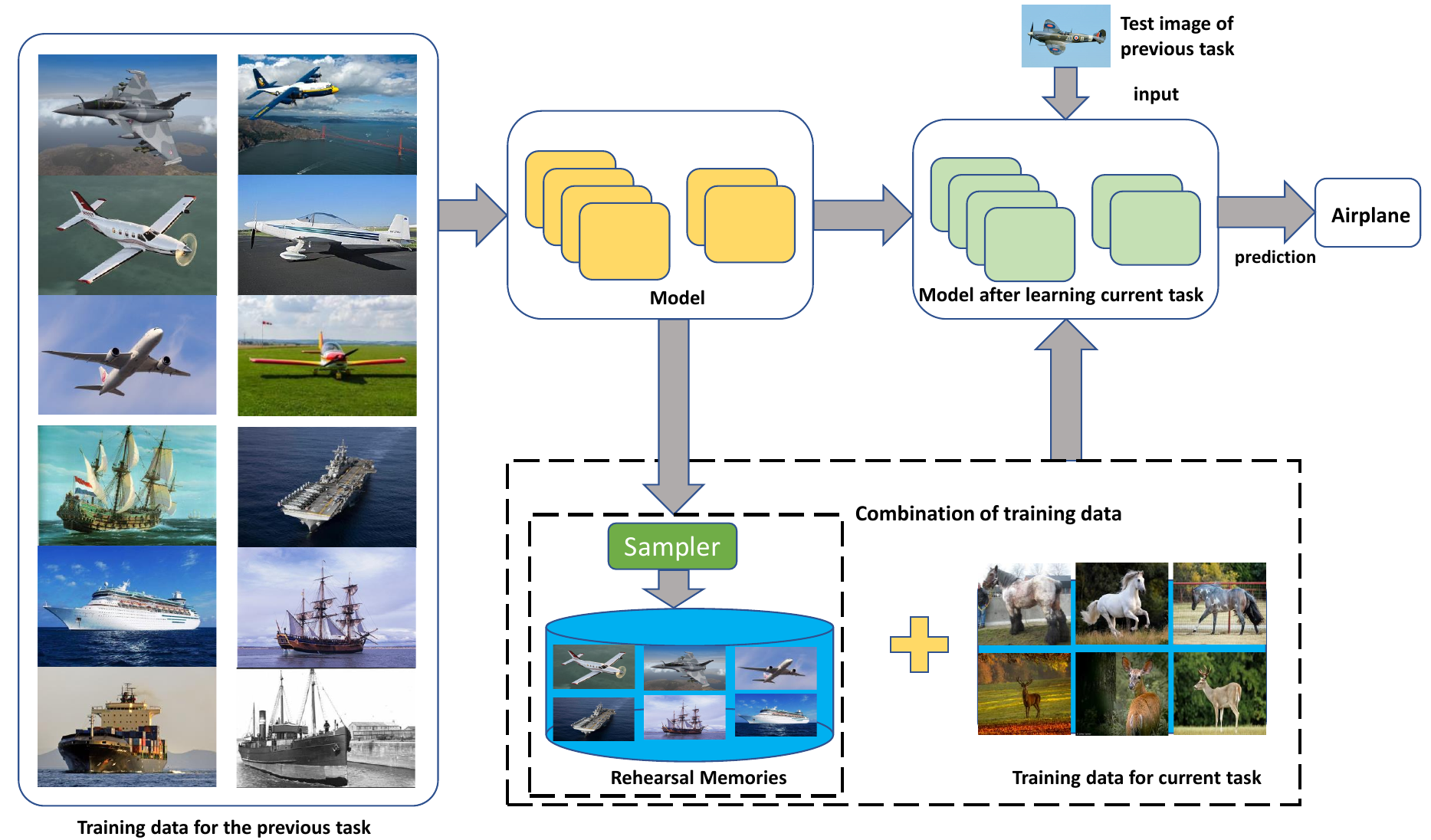}
    \caption{Architecture of proposed rehearsal-based CL}
\end{figure*}

    We outline the core components of a generic rehearsal-based class-incremental learning system.
    
 \begin{itemize}
      \item \emph{Representative memory}: Representative memory is used to store and manage data points that best represent what the model has learned in previous tasks. Storing all prior data points may be impracticable due to storage issues. 

      \item \emph{Model}: A general rehearsal-based architecture employs a deep neural network that has been trained on two different loss functions, such as distillation loss and cross-entropy loss, and it is typically called a cross-distilled loss function. Generally, a ResNet model is used as a trainable feature extractor.
      
      \item \emph{Weighting Policies}:
      In real-world CL systems, some tasks may have more importance than others, whereas, during training, distribution techniques fail to take into account the significance of each task \cite{nokhwal2023embau, nokhwal2024survey, nokhwal2024hffa, nokhwal2024iiot, nokhwal2024gsp, nokhwal2024opt}. Thus, it is important to consider all tasks with some weighting policy.

      \item \emph{Classifier}:
      CL's inference setup can be mainly divided into two categories: single-head  configuration and multi-head configuration \cite{chaudhry2018riemannian}.
      When using single-head configurations, there is only one head (single-classifier), hence no task ID is required during inference.
      For example, iCaRL \cite{rebuffi2017icarl} uses an NME (nearest-mean-of-exemplars) classifier based on the mean of the stored exemplars in the feature space. 

      \end{itemize}

\section{Proposed Technique}
\seclab{proposed_tech}
We present an exemplars management technique that preserves the boundaries of a class distribution by selecting a varied set of data points to deal with the disjoint and fuzzy CIL configurations. Such a varied set is particularly necessary in the fuzzy CIL scenario, as classes can repeat across tasks. The exemplars that the sampling algorithm chooses to keep in memory are representative of their class.

\textbf{\emph{Sampling algorithm:}}
This sampling algorithm is inspired by the $2$-approximation algorithm for metric $K$-center clustering algorithm \cite{gonzalezKCC}. 
Initially, the higher-dimensional training dataset $D_t$ for task $t$ is reduced to a lower-dimensional space, such as 2D using t-SNE \cite{van2008visualizing}. For our experiments, an arbitrarily chosen 2 is used as this is computationally very efficient and without losing the important structure of data. However, $X$ can also be reduced to slightly higher dimensions.

Then, the closest point to the mean is selected and stored in the rehearsal memory denoted as $P$. Subsequently, another exemplar $e$ is chosen, which is furthest from the current set of exemplars $P$, and has at least $n$ neighboring data points within a distance $r$ ($n, r$ are supplied as hyper-parameters). This is continued until $m$ points are selected. The pseudocode for our sampling algorithm is presented in Algorithm~\ref{algo:sampler}.

The choices for $n, r$ can be adjusted to get the best results. For a reasonable $n$, and very large $r$, the algorithm will produce the usual $2$-center approximation to $K$-center clustering with $K=m$. A very small value of $r$ may lead to closely chosen centers if there are only a few points in a tight cluster and the rest are spread out. Also, a very small value of $r$ or a very large value of $n$ may cause no point to be available during an iteration - if this happens, the algorithm is run again by adjusting $r$ or $n$ or both. For example, $r$ can be increased by some small $\delta > 0$ or $n$ can be decreased by some constant amount. In our experiments, we set $r$ to an arbitrary value of 0.5 and increase it by 0.1 every time a new potential exemplar does not satisfy the criteria.

\textbf{\emph{Training:}} The overall training of our approach is inspired from \cite{rebuffi2017icarl}. For the first task, only the cross-entropy loss is used, and its contribution by an arbitrary training point $x$ can be mathematically formulated as:
\begin{equation}\label{eq:crossLoss}
    \mathcal{L}_{CE}(x) = \sum_{i = 1}^{\ell + q} -\hat{y}^{(i)}(x) \log\left[p^{(i)}(x)\right]
\end{equation}
Here, $ \hat{y}(x) = (\hat{y}^{(1)}(x), \ldots, \hat{y}^{(\ell+q)}(x))$ is the predicted one-hot encoding for $x$, $\ell$ is number of training classes which are already experienced (until $(t-1)$ tasks) by the model, and $q$ is the number of classes in the current task $t$. The logits predicted for $x$ are $p(x) = (p^{(1)}(x), p^{(2)}(x), \ldots, p^{(\ell+q)}(x))$.
Now, the knowledge distillation loss (for $x$) can be formulated as:
\begin{equation}\label{eq:kdLoss}
    \mathcal{L}_D(x) = \sum_{i = 1}^{\ell} -\hat{p}_{T}^{(i)}(x) \log\left[p_{T}^{(i)}(x)\right]
\end{equation}
The logits for the teacher model are $\hat{p}(x) = (\hat{p}^{(1)}(x), \hat{p}^{(2)}(x), \ldots,\hat{p}^{(\ell)}(x))$, and $\hat{p}_{T}^{(i)}(x)$ and $p_{T}^{(i)}(x)$ are the $i^{th}$ distilled output logit and can be written as:
\begin{align*}
\hat{p}_T^{(i)}(x) &= \dfrac{\exp(\hat{p}^{(i)}(x)/T)}{\sum_{j=1}^{\ell}\exp(\hat{p}^{(j)}(x)/T)}, \\ 
p_T^{(i)}(x) &= \dfrac{\exp(\hat{p}^{(i)}(x)/T)}
{\sum_{j=1}^{\ell}\exp(p^{(j)}(x)/T)}
\end{align*}
The temperature scalar  $T>1$ makes the data distribution smoother. Summing equations \ref{eq:crossLoss} and \ref{eq:kdLoss}, the final cross-distillation loss contribution for $x$ is,
\begin{equation}\label{eq:finalLoss}
    \mathcal{L}_{CD}(x) \\
    = \beta \mathcal{L}_D(x) + (1-\beta) \mathcal{L}_{CE}(x)
\end{equation}
A hyper-parameter $\beta$ is used to fine-tune the effect of equations \ref{eq:crossLoss} and \ref{eq:kdLoss}; $\beta$ was set to 0.5 for our experiments. Finally, for the first task, the loss function is,
\[
\mathcal{L} = \sum_{x \in \mathcal{D}_1} \mathcal{L}_{CE}(x),
\]
where $\mathcal{D}_1$ denotes the data in the first task and, 
\[
    \mathcal{L} = \sum_{x \in \mathcal{D}_j \cup P_j} \mathcal{L}_{CD}(x),
\]
for task number $j \geq 2$, where $\mathcal{D}_j$ denotes data for the $j$th task and $P_j$ is the exemplars set at the $j$th task.

\begin{algorithm}[ht]
    \DontPrintSemicolon
    \SetKwFunction{FSampler}{DSS\_Sampler}
    \SetKwProg{Fn}{Function}{:}{}
    \Fn{\FSampler{$X, m, n, r$}}{
        \KwIn{Image training set $X = \{x_1, \mathellipsis, x_n\}$ of class $c$, $m$ -- number of output exemplars, $n$ -- the number of data points within a distance of $r$ (used to exclude outliers)}
        \KwOut{ a set of $P$ exemplars}

        \BlankLine
        $R_D \gets$ Reduce the dimensionality of $X$ to 2 using t-SNE \cite{van2008visualizing}    \\    
        $P \leftarrow \{e\}$ where $e$ is the closest point to the mean chosen from $R_D$\\
            \While{$|P| < m$}{
                Order points of $R_D$ in reverse by distance $d_P[x]$ from $P$;
                $d_P[x] = \min_{y \in P} d(x,y)$ \\
                
                $e \gets$ Pick first (furthest from $P$) point in ordering such that it has at least $n$ points within distance $r$ \\

                $P \gets P \cup \{e\}$ \\
            }
        \Return points from $X$ corresponding to $P$ ($\subseteq R_D$)
    }
\caption{Algorithm for exemplars sampling}
\label{algo:sampler}
\end{algorithm}

        



\section{Experiments}
\seclab{experiments}
\begin{figure*}[htbp!]
    \centering    
    \includegraphics[height=5.3cm]{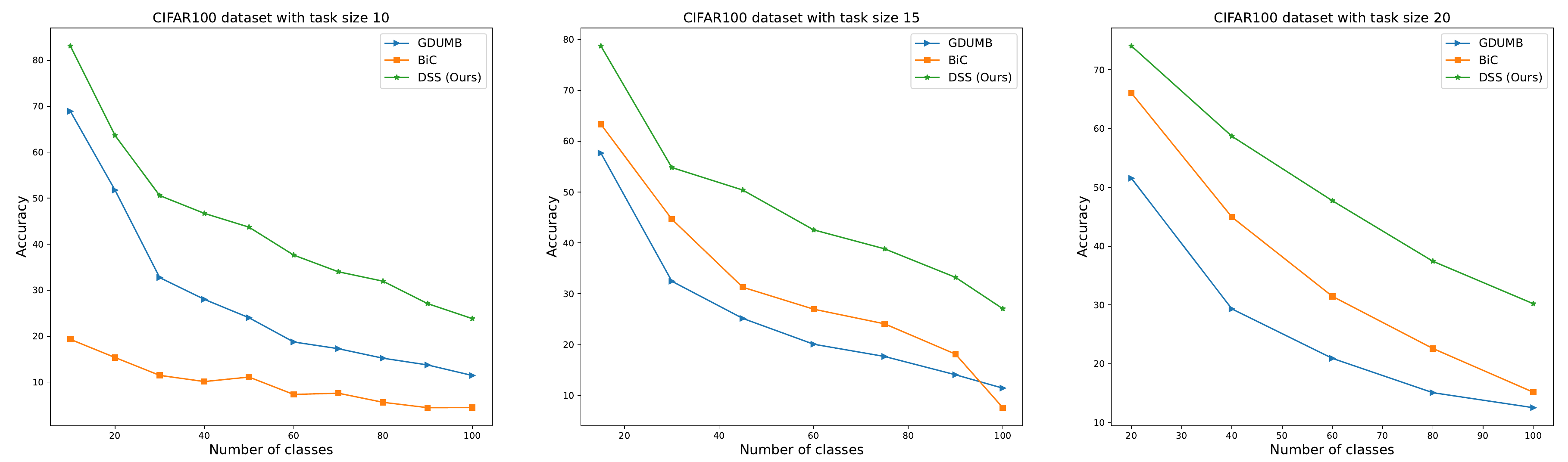}
    \caption{Comparative results of state-of-the-art methods using `'Disjoint'' setup}
    \label{fig:results_disjoint}
\end{figure*}

\begin{figure*}[htbp!]
    \centering    
    \includegraphics[height=5.3cm]{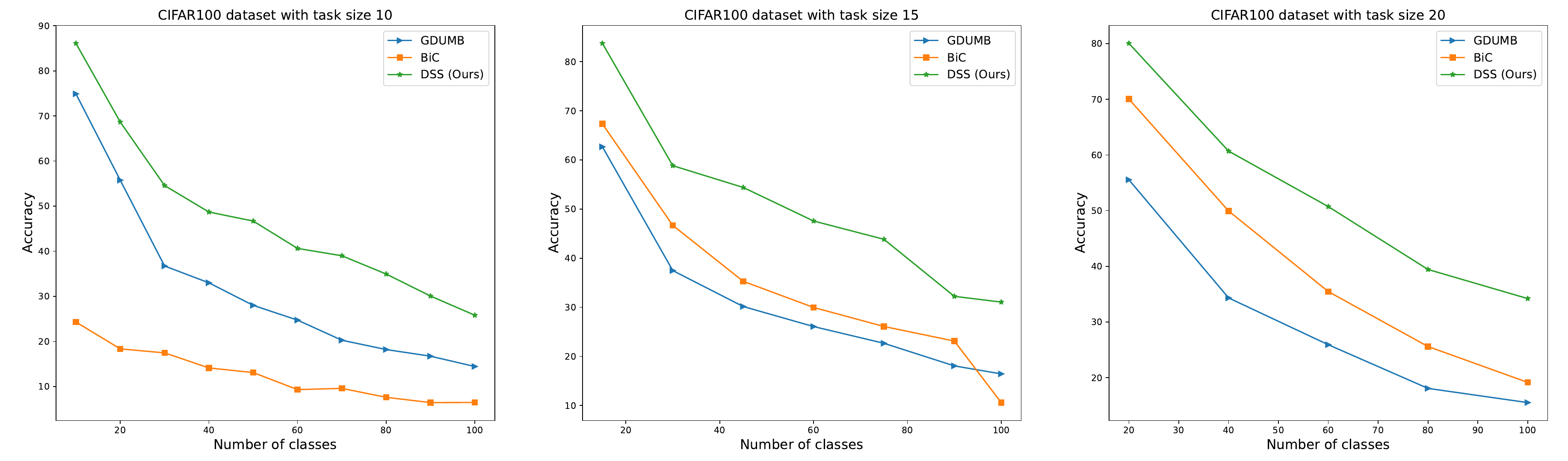}
    \caption{Comparative results of state-of-the-art methods using `'Fuzzy10'' setup}
    \label{fig:results_fuzzy10}
\end{figure*}

By contrasting DSS to other state-of-the-art techniques in various experimental settings, i.e., CIL task configurations acting as benchmarks, and different values of $n$ neighboring data points, we provide empirical validation of DSS's effectiveness.

\subsection{Dataset and Metrics}
We have used the\datasetOne dataset for our experiments. It includes 60,000 RGB images with a resolution of 32x32 pixels, organized into 100 categories with 6,000 images in each category. For training and testing, 50,000 and 10,000 images are used, respectively. For the graphs, we reported the results using accuracy for each task, and  "Average Accuracy"(AA) is used for studying the effects of $n$ neighboring points. AA refers to the overall performance score achieved by averaging the results of all tasks.


\subsection{Experimental setup}
Unless otherwise mentioned, all of our algorithm evaluations are conducted in disjoint-CIL and fuzzy10 scenarios. Experiments for\datasetOne have been conducted using ResNet32 architecture. The ReLu activation function was applied with a batch size of 128 samples \cite{tanwer2020system}. For\datasetOne dataset, the memory capacity of the exemplars was set to 1000 (representing less than 2\% of the training samples of each class). The total number of epochs was 70. We used 0.01\% for both learning rate and momentum.

\subsection{Benchmark Task Setup}
We evaluate our proposed DSS method against well-established CIL strategies, such as BiC \cite{wu2019large} and GDumb \cite{prabhu2020gdumb}. All of the tests were performed in disjoint-offline and fuzzy-offline setups. With an exemplar's memory, a model may utilize the incoming data repeatedly (i.e. epochs) in an offline configuration. In a CIL scenario, the inference does not have the task-ID. The use of a single-head classifier in conjunction with a joint final layer has been provided for each task.

\subsection{Results}
Figures \ref{fig:results_disjoint} and \ref{fig:results_fuzzy10} highlight the results of our comparison of the developed DSS to state-of-the-art approaches in disjoint-offline and fuzzy10-offline scenarios respectively. The graph demonstrates that DSS outperforms BiC and GDumb, consistently by at least 10\% in accuracy.

\subsection{Influence of '$n$' neighbor data points on average accuracy} 
In our ``DSS'' approach, the influence of outliers decreases as $n$ grows, resulting in a more stable and consistent classification. Since outliers may be near the potential exemplar point when $n$ is small, their effect on the predictions might be exaggerated as shown in the experimental results. As $n$ increases, nevertheless, the impact of any particular outlier diminishes, since a large $n$ ensures chosen exemplars are surrounded by a larger number of other data points and are therefore most likely not an outlier. As a result, the algorithm is better equipped to achieve reliable results even when exposed to outlier data points and abnormalities.


\begin{table}[ht]
\caption{Influence of $n$ neighbor size on average accuracy for (a) Disjoint setup and (b) Fuzzy10 setup}
\centering

\begin{subtable}[t]{0.45\linewidth} 
    \caption{}
    \centering
    \begin{tabular}{|c|c|c|c|}
    \hline
        $n$=0 & $n$=3 & $n$=5 & $n$=8 \\ \hline
        49.62 & 50.2 & 51.58 & 51.93 \\ \hline
    \end{tabular}
    \label{tbl:neighbor_effect_tbl1}
\end{subtable}

\vspace{1em} 

\begin{subtable}[t]{0.45\linewidth} 
    \caption{}
    \centering
    \begin{tabular}{|c|c|c|c|}
    \hline
    $n$=0 & $n$=3 & $n$=5 & $n$=8 \\ \hline
    53.02 & 54.34 & 54.67 & 54.8 \\
    \hline
    \end{tabular}
    \label{tbl:neighbor_effect_tbl2}
\end{subtable}
\end{table}

\subsection{Intuition behind our sampling algorithm}
While working in the high-dimensional space, t-SNE concentrates on maintaining the data's local structure by emphasizing the shared characteristics between encompassing points. As a result, t-SNE is better able to detect subtle clusters and structures that weren't immediately obvious in the input data. Thus, t-SNE is robust in the face of outliers. 

The basic premise of the technique is to choose those exemplars that best represent the data over a series of iterations, to maximize the distance between each exemplar, having a minimum of $n$ neighbors within $r$ distance apart. Using this method, the algorithm chooses representative data points (exemplars) without including any outliers, which would otherwise skew the results. The filtering process used to accomplish this resilience takes into account the proximity of each potential data point to its neighbors. This method assures that the selected data point is consistent with its surrounding data and not an anomaly.

\section{Conclusion}
\seclab{conclusion}
We presented a novel sampling technique (``DSS'') based on $K$-center clustering that is simple, efficient to implement, ensures diversity of stored exemplars, and is robust to data outliers, in offline class incremental learning scenarios for both fuzzy and disjoint task boundaries. Our method outperforms state-of-the-art methods in such settings.

\bibliographystyle{ieeetr}
\bibliography{bibliography}

\end{document}